# From Data to Action: Exploring AI and IoT-driven Solutions for Smarter Cities

Tiago Dias[12[0000-0002-1693-7872]], Tiago Fonseca[1[0000-0002-5592-3107]], João Vitorino[12[0000-0002-4968-3653]], Andreia Martins[12[0000-0001-8017-7460]], Sofia Malpique[12[0000-0003-2214-7495]] and Isabel Praça[12[0000-0002-2519-9859]]

[1] School of Engineering, Polytechnic of Porto (ISEP/IPP), 4249-015 Porto, Portugal
[2] Research Group on Intelligent Engineering and Computing for Advanced Innovation and Development (GECAD), 4249-015 Porto, Portugal
`{tiada,calof,jpmvo,teles,pique,icp}@isep.ipp.pt`

**Abstract.** The emergence of smart cities demands harnessing advanced technologies like the Internet of Things (IoT) and Artificial Intelligence (AI) and promises to unlock cities' potential to become more sustainable, efficient, and ultimately livable for their inhabitants. This work introduces an intelligent city management system that provides a data-driven approach to three use cases: (i) analyze traffic information to reduce the risk of traffic collisions and improve driver and pedestrian safety, (ii) identify when and where energy consumption can be reduced to improve cost savings, and (iii) detect maintenance issues like potholes in the city's roads and sidewalks, as well as the beginning of hazards like floods and fires. A case study in Aveiro City demonstrates the system's effectiveness in generating actionable insights that enhance security, energy efficiency, and sustainability, while highlighting the potential of AI and IoT-driven solutions for smart city development.

## 1   Introduction

The growth and urbanization of cities worldwide, associated with increasing population expectations, has led to a multitude of complex challenges across various domains, such as transportation, public safety, energy consumption, and infrastructure maintenance and management. These challenges can result in significant negative impacts on both the environment and citizens' standard of living, making them into a compelling impetus for the efforts of city planners, policymakers, engineers, and the public at large. Against this backdrop, the emergence of smart cities, powered by the use of advanced Information and Communication Technologies (ICT), such as the Internet of Things (IoT), Artificial Intelligence (AI), Machine Learning (ML), and data analytics, promise to unlock cities' potential to become more sustainable, intelligent, efficient, and ultimately livable for their inhabitants.

Despite its novelty and innovative nature, several cities around the world have already begun to implement smart city solutions [1]. For instance, Barcelona, Spain, has established a comprehensive smart city platform [2] that includes various solutions, such as smart parking, waste management, and air quality monitoring, all aimed at improving urban life for citizens. In Singapore [3], the Smart Nation initiative aims to harness technology to improve the quality of life of residents by enhancing transportation, healthcare, and public safety. In Portugal, the city of Aveiro is one of the city's leading these innovations and has made significant investment towards becoming a smarter city. The city has established the Aveiro Tech City Living Lab (ATCLL) [4], which is a research platform that serves as a testing ground for smart city solutions.

This work introduces the Intelligent City Management System (ICMS), the result of our participation in the first edition of the Aveiro Tech City hackathon, promoted by the municipality of Aveiro, that aimed to further enhance the capabilities and services of the city's management platform. ICMS is designed to enhance city management through a scalable and intuitive AI-powered system that integrates multiple data analysis and prediction dashboards. The system impact is evaluated by a live case study using real-world data derived from the ATCLL environment and additional IoT sensors available during the hackathon.

## 2    State-of-the-Art

The concept of a "smart city" has emerged as a response to the economic, social, and political challenges faced by post-industrial societies at the outset of the new millennium. This idea involves the strategic use of digital technologies to develop innovative solutions for urban communities. Therefore, the primary objective is to address the challenges encountered by urban society, including environmental pollution, demographic change, healthcare, the financial crisis, and resource scarcity [5].

Novel advancements in IoT are a key enabler for smart city applications [6], being responsible for generating an enormous quantity of data [7]. Indeed, in [8] Allam et al., have put forth a novel Smart City framework that integrates AI technology and urban systems, with a primary focus on enhancing urban sustainability and liveability. The authors contend that technology ought to serve as a fundamental cornerstone of Smart Cities, where Big Data can be derived from various domains through IoT. AI is proposed as an underlying feature capable of processing, analyzing, and interpreting the generated data. Moreover, to evaluate the electricity consumption patterns in Iran, Ghadamiet al. employed machine learning techniques and implemented dynamic strategies to foster citizen participation in renewable energy generation, informed by expert knowledge. The authors utilized a combination of an Artificial Neural Network and statistical analysis to develop a Decision Support System [9].

Another IoT applications for smart cities are related to vehicular traffic data, which represents one of the most vital data sources in a typical smart city. Effective analysis of this data can yield significant benefits for both citizens and governments. Neyestani et al. proposed a Mixed-Integer Linear Programming model for the traffic behavior of Plug-in Electric Vehicles, which can be integrated as a sub-module in various other studies such as operation and planning, thereby providing decision makers with valuable insights in urban environments [10], [11].

In this context, the ATCLL an open platform for developing, testing, and demonstrating innovative concepts, products, and services related to the urban environment. It includes an advanced communication infrastructure and an urban data management and analytics platform that can collect, process, and analyze data from various sources. The platform offers opportunities for anyone, or any organizations interested in devising novel solutions for the predicaments encountered in contemporary urban settings [12].

ATCLL integrates a communication infrastructure and sensing platform, which comprises an array of smart lamp posts that are equipped with both traffic radars and video cameras. Additionally, the platform integrates buses and other vehicles that are fitted with devices that collect and transmit data. Furthermore, sensors are deployed throughout the city to monitor the number of people present in different zones, as well as to measure environmental quality and other relevant factors. The seamless integration of these components creates a comprehensive and sophisticated technological ecosystem that enables the collection and analysis of vast amounts of data, providing new and innovative ways to address the challenges faced by modern cities. Overall, the ATCLL is a cutting-edge initiative that combines technology, research, and innovation to create a living laboratory for urban development [4], [12].

## 3  Proposed Solution

The literature shows that smarter cities generate an enormous flow of information which can be useful to keep track, improve and solve issues inherent to the city. Ultimately, its progression and development provide its inhabitants with better life quality. However, infrastructure costs, privacy and security issues and interoperability of multiple systems can be an embargo to achieve this goal. As such, this work attempts to facilitate and leverage the implementation of smart devices installed across Aveiro city to create an Intelligent City Management System (ICMS).

The proposed system is an AI-powered comprehensive system that integrates multiple data analysis and prediction dashboards to provide a single point of management for the city. The system provides a holistic view of various sectors of the city, with analytics and forecasting capabilities that allow city managers to make decisions quickly and effectively, improving efficiency and resource allocation. To enable its use across different cities, the system is highly scalable and configurable.

The proposed system is divided into four different components: (i) City Security and Safety (CSS), (ii) City Energy Management (CEM), (iii) City Infrastructure Maintenance (CIM), and (iv) City Management Dashboard (CMD). In this Representational State Transfer (REST) architecture, each component is considered a REST API and their communication is based on HTTP request. The authors decided to follow this architecture, as it allows for very low coupling of the components that represent different city management fields, making the system highly scalable, reliable, portable, and maintainable (Fig. 1).

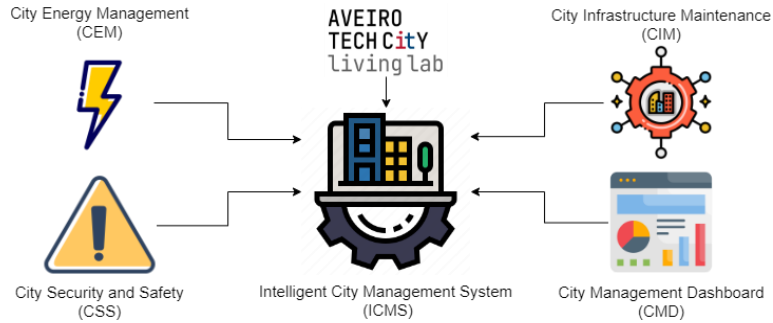

**Fig. 1.** ICMS architecture.

Each component is implemented in Python and provides a real-time analysis of several aspects of the city, using data gathered from the ATCLL. The following subchapters define the problem, the goal, and the implementation strategy of each city management component.

### 3.1 Use Case 1: Security and Safety

Ensuring the security and safety of both vehicles and pedestrians is of paramount importance in modern cities, as it not only protects the well-being of individuals but also contributes to the overall livability and sustainability of the city.

The CSS component focuses on improving road and pedestrian safety, by analyzing data provided by smart posts spread across Aveiro city. As described in Section 2, these smart posts are equipped with multiple sensors, of which cameras and speed radars, that are strategically placed to capture pedestrian and vehicle circulation in the same area. The authors considered the use of this information relevant to monitor in real-time driver and pedestrian safety. The premise of this correlation is that a driver should adapt his/her driving behavior depending on the number of pedestrians in a certain zone, since the probability of an accident that compromises safety of those around is much higher. The goal of this integrated component is to provide intelligently organized data and assist on the decision-making process of security implementations regarding the city's public highways. CSS works similarly to an expert system, as it is capable of correlating information using user defined rules, however, it further expands on it by being capable of performing feature computation. These features and rules should be managed by the city's security decision-makers.

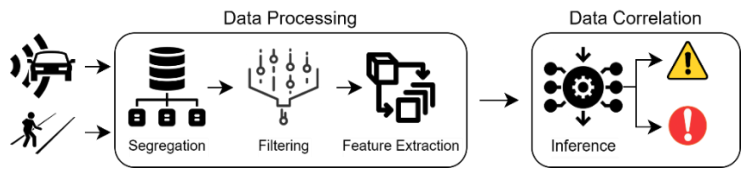

**Fig. 2.** CSS pipeline overview.

5As described in Fig. 2, this use case is divided into the data processing and data correlation phases. The firstly the data consumed is segregated by smart post, to ensure correctness of the correlations. Then, the radar data that is unrelated to heavy or light vehicles are discarded. Lastly, for each smart post, the speed average and the number of pedestrians features are computed for the same time frame, which corresponds to the cadency configured, to aggregate the occurrences in each zone.

The second phase takes the processed intelligible data and correlates it according to rules defined by the decision-makers. The resulting correlation is then classified by the rules as warning or danger depending on the severity of the violation. Since the data that is correlated always belong to the same place, the city's security decision-makers can visualize where the violations are occurring and take into consideration a frequency level that is presented to decide whether security measures should be taken.

Even though this implementation only considers information regarding radar sensors and pedestrian count, other information can be included in the rules, as long as it is captured by the smart posts. The addition of other information, such as the existence of walkways or not can be included to attain a more fine-grained security analysis between pedestrians and vehicles.

### 3.2   Use Case 2: Energy Management

The global environmental changes and ongoing energy crisis have amplified the need for efficient energy management in urban environments. As cities around the world strive to become smarter cities, they are actively exploring ways to optimize their energy consumption and reduce their carbon footprints. In this context, City Energy Management (CEM) has emerged as a crucial component in the design and operation of our smart city platform (Fig. 3).

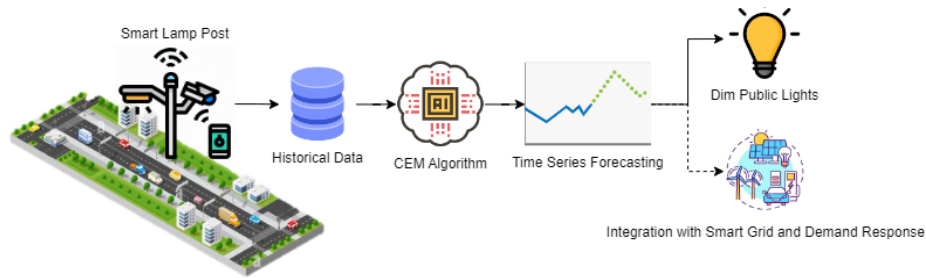

**Fig. 3.** CEM pipeline overview.

Our solution utilizes the ATCLL smart lamp posts, which are equipped with a variety of sensors and cameras. Given their historical data on the number of identified pedestrians, vehicles, and other moving objects in each street, our algorithm is designed to predict the number of movements likely to occur on the street in the next 24 hours, accounting for the differences between workdays and weekends. Based on these predictions, CEM can provide recommendations on when to dim public lighting in specific streets. Alternatively, if the lights do not support dimming, CEM could advise shutting off half of the lamp posts in the street. This approach to public lighting enables a city to reduce its energy consumption while maintaining public safety.

Moreover, we highlight the possibility of integrating CEM with smart energy communities and intelligent demand response strategies, such as [13]. This can bring synergistic advantages because utility providers can effectively optimize and schedule flexible energy resources and energy storage across the city, leading to a reduction in costs and peak demand. By forecasting the absence of people on several streets at night, the system can dim the public lights in an event of peak grid consumption, acting as a smart regulating reserve mechanism. Participation in such mechanisms can even generate revenue for the city. Consequently, CEM not only acts as an isolated smart-city solution, but it can also be part of the creation of a more efficient and robust energy grid for urban areas, while incentivizing the use of renewable energy.

Regarding the specifically designed AI time-series forecasting algorithm, its implementation process consisted of four steps: collection and preprocessing of historical data, feature engineering, model selection, and model training. These are the steps that permit our algorithm to learn the patterns and relationships between various factors that influence the number of pedestrians, vehicles, and objects on the streets of a city. First in the data collection and preprocessing step, we collect historical data from the smart posts in Aveiro, which includes pedestrian and vehicle counts, as well as weather conditions, day of the week, holidays, time of day, and local events. The data is then preprocessed to remove any outliers, missing values, and inconsistencies.

Next, during the feature engineering step, we extract relevant features from the preprocessed data, which include temporal features (e.g., hour of the day, day of the week), weather features (e.g., temperature, humidity), and event-based features (e.g., holidays, local events). These features are crucial for improving the accuracy of our model. Finally, we selected and trained our machine learning model for time series forecasting using the preprocessed data and features, adjusting hyperparameters as necessary to minimize the error in the forecasts.

### 3.3 Use Case 3: Infrastructure Maintenance

A city's infrastructure is essential for its efficient functioning, and regular maintenance is crucial to ensure its longevity. However, regular wear and tear and other factors can cause these infrastructures to fail, often leading to costly maintenance issues. Therefore, monitoring the infrastructure of a city is crucial to its development but the identification of maintenance issues can be challenging and time-consuming.

As part of the ICMS platform, CIM (Fig. 4).attempts to automate and improve the monitorization of Aveiro by leveraging its smart public transportation to efficiently monitor in a distributed way the city's infrastructure, resorting to live-image capturing and computer vision to detect infrastructure defects, which in turn are reported in real-time to the city's infrastructure engineers, allowing them to make data-driven decisions to ensure maintenance of the infrastructure.

The You Only Look Once (YOLOv5) [14] algorithm is utilized within the component to perform object detection of maintenance issues and the beginning of hazards, using a live feed by the smart public transportation. The algorithm was trained using three annotated datasets, the Pothole Object Detection Dataset [15], the Roadway Flooding Image Dataset [16] and the FIRE Dataset [17], to detect the occurrence of potholes, floods, and fires, which are three concerning aspects for the city of Aveiro.

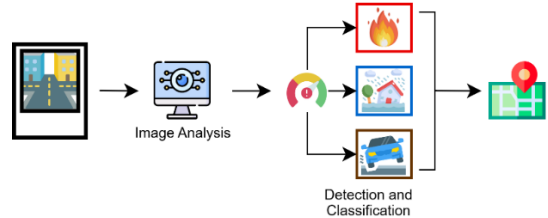

**Fig. 4.** CIM pipeline overview.

The CIM execution pipeline consists of analysing the captured images of the city relying on the employed YOLOv5 model. The algorithm detects the city's infrastructure defects, highlighting them with bounding boxes and assigning them a confidence score between 0 and 1. A higher confidence score reflects worse conditioning of the infrastructure and therefore requires more immediate attention. Lastly, the coordinates of detected issues are presented along with the highlighted images to the city's infrastructure engineers in an interactive map, so that they can be remotely analysed to decide which actions should be taken.

## 4  Case Study

An empirical case study was carried out to assess the feasibility and reliability of the proposed solution for the city of Aveiro. The organizing committee of the Aveiro Tech City hackathon provided two months of recorded data from the ATCLL, so the first month could be used for training of ML models, and the second for a holdout evaluation. ICMS was calibrated to the characteristics of the existing smart infrastructure and fine-tuned to the data readings of the city's IoT sensors. Then, the capabilities of the system were demonstrated live in the final stage of the hackathon.

Regarding the first use case, Security and Safety, the ratio between the number of speeding vehicles and the number of pedestrians, per hour of the day, can be visualized for each street of Aveiro equipped with the smart infrastructure. For the analyzed month, the ratio exceeded the allowed threshold in several streets and therefore further speed reduction mechanisms like speed humps and rumble strips are required to compel drivers to reduce speed. For instance, this can be noticed in a street where the threshold was slightly exceeded on a weekly basis (Fig. 5). Additionally, in this street and some nearby streets, there was a significant spike by the end of the month, possibly due to an event occurring in that area of the city with many pedestrians and vehicles in circulation. It could be valuable to correlate the sensor data with information about ongoing events in the city, to better distinguish between sporadic spikes and areas where drivers exbibit dangerous behavior on a regular basis.

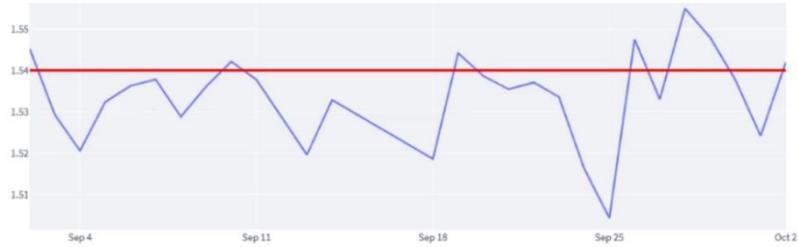

**Fig. 5.** CSS threshold analysis.

Regarding the second use case, Energy Management, the number of pedestrians, vehicles, and other moving objects can be visualized for each street. To distinguish between day and night times, the latter have a darker grey background. Furthermore, the hours when no activity was registered on a street are highlighted in green blocks to indicate that power consumption could be reduced in those hours by shutting off public equipment or dimming public lighting. For instance, these blocks where efficiency could be improved can be noticed almost every night in Aveiro (Fig. 6).

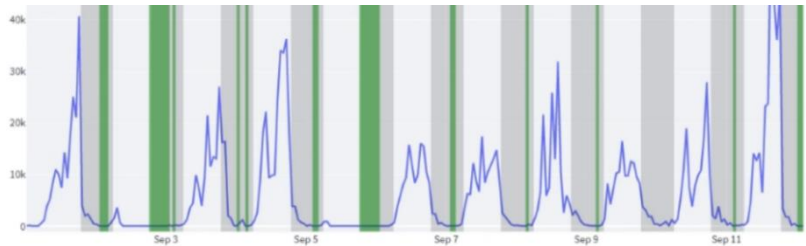

**Fig. 6.** CEM block identification.

Additionally, based on the historical data provided in the first month, the proposed solution can predict the number of movements likely to occur on the street in the next 24 hours at a time, enabling a forecasting of the best hours to apply energy saving measures (Fig. 7). These predictions achieved a good generalization throughout the second month of data, which demonstrates that cost savings could be achieved with more intelligent management of public lighting. If such algorithms were trained with an entire year of data, they could be improved to account for special holidays and events that may affect the hours of activity in different areas of the city.

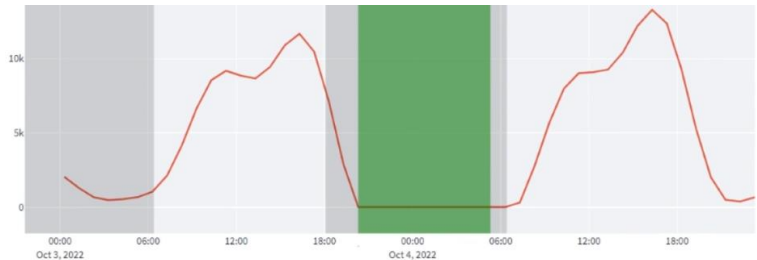

**Fig. 7.** CEM activity forecasting.

Regarding the third use case, Infrastructure Maintenance, every maintenance issue was created as an occurrence with location coordinates to be displayed in the interactive map of the ATCLL. For instance, in one of the main roundabouts of the city, a pothole was detected in a vehicle that was simulating the route of a bus, checking if it could be used as a mobile camera platform. The pothole was automatically assigned a confidence score of 0.41, which indicates that it is not an urgent issue, but it is still relevant for the municipality to fix it (Fig. 8). It is pertinent to note that the live feed of the camera was analyzed in real-time and immediately discarded afterwards, so only the frames where a public maintenance issue was detected were stored. Further cybersecurity measures like the anonymization of license plates are essential to comply with privacy regulations in smart city solutions that rely on camera feeds.

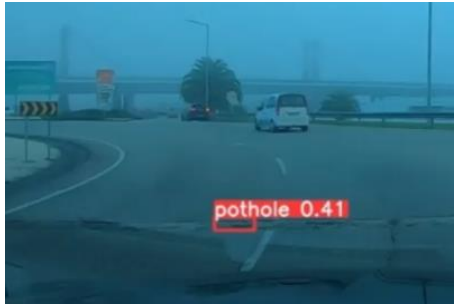

**Fig. 8.** CIM issue detection.

## 5      Conclusion

This work addressed several possible applications of AI to smart cities, in the context of the first edition of the Aveiro Tech City hackathon. The proposed system, ICMS, provides a data-driven approach to three use cases: (i) analyze traffic information to reduce the risk of traffic collisions and improve driver and pedestrian safety, (ii) identify when and where energy consumption can be reduced to improve cost savings, and (iii) detect maintenance issues like potholes in the city's roads and sidewalks, as well as the beginning of hazards like floods and fires.

By harnessing the power of AI and IoT, the proposed system can be significantly beneficial to the security, energy efficiency, and sustainability of a smart city. Further research efforts must be made to develop smart city solutions capable of tackling the environmental challenges of urban environments, so cities like Aveiro can provide more security and a better quality of life to their citizens.

**Acknowledgements.** The authors would like to thank the University of Aveiro, Instituto de Telecomunicações and Câmara Municipal de Aveiro for organizing the event and proving the city data utilized in this work.

This work has received funding from UIDB/00760/2020 and from UIDP/00760/2020.